# PrimSeq: a deep learning-based pipeline to quantitate rehabilitation training

Avinash Parnandi[1,7], Aakash Kaku[2,7], Anita Venkatesan[1], Natasha Pandit[1], Audre Wirtanen[1], Haresh Rajamohan[2], Kannan Venkataramanan[2], Dawn Nilsen[3], Carlos Fernandez-Granda[2,5,8], Heidi Schambra[1,5,6,8]

**Affiliations:**

[1] Department of Neurology, New York University Langone Health, New York, USA
[2] Center for Data Science, New York University, New York, USA
[3] Department of Rehabilitation and Regenerative Medicine, Columbia University, New York, USA
[4] Courant Institute of Mathematical Sciences, New York University, New York, USA
[5] Department of Rehabilitation Medicine, New York University Langone Health, New York, USA
[6] Neuroscience Institute, New York University Langone Health, New York, USA
[7] These authors contributed equally: Avinash Parnandi, Aakash Kaku.
[8] These authors jointly supervised this work: Carlos Fernandez-Granda, Heidi Schambra.
Corresponding author emails: cfgranda@cims.nyu.edu; heidi.schambra@nyulangone.org.

**Abstract**

Stroke rehabilitation seeks to increase neuroplasticity through the repeated practice of functional motions, but may have minimal impact on recovery because of insufficient repetitions. The optimal training content and quantity are currently unknown because no practical tools exist to measure them. Here, we present PrimSeq, a pipeline to classify and count functional motions trained in stroke rehabilitation. Our approach integrates wearable sensors to capture upper-body motion, a deep learning model to predict motion sequences, and an algorithm to tally motions. The trained model accurately decomposes rehabilitation activities into component functional motions, outperforming competitive machine learning methods. PrimSeq furthermore quantifies these motions at a fraction of the time and labor costs of human experts. We demonstrate the capabilities of PrimSeq in previously unseen stroke patients with a range of upper extremity motor impairment. We expect that these advances will support the rigorous measurement required for quantitative dosing trials in stroke rehabilitation.

**Introduction**

Most individuals with stroke have persistent motor deficits in an upper extremity (UE)[1]. Rehabilitation training is meant to promote UE motor recovery, and patients practice functional activities that consist of object-directed UE motions[2]. In animal models of rehabilitation, UE recovery is potentiated by large quantities of functional motions trained early after stroke[3-5]. In humans, the relationship between recovery and quantity of UE functional motions is less clear, primarily because an effective method to count these motions does not currently exist. The quantification of functional motion repetitions—a critical parameter in rehabilitation dose[6]—has been challenged by the trade-off between pragmatism and precision.

Most clinical rehabilitation studies opt for pragmatism, using time in therapy to estimate how much training has occurred[7-13]. Problematically, time metrics do not directly read out motion content or quantity. The number of repetitions vary considerably in a given therapy time, and functional motions may not even be practiced[14, 15]. Investigators have advanced the automated distinction of functional versus nonfunctional UE motion[16-18], but this approach still measures time spent in motion, rather than *what* and *how many* functional motions were performed. The imprecision of time metrics in rehabilitation research hampers the ability to establish dosing parameters that meaningfully impact recovery. This lack of granularity also undermines the delivery of evidence-based rehabilitation, as replicating time in therapy does not ensure replication of the repetition quantity that actually drives recovery.

Some rehabilitation studies have instead opted for precision by manually counting functional motion repetitions[15, 19-22]. This approach meticulously identifies training content and quantity, but has practical challenges. Functional motions are fluid, fast, and difficult to disambiguate in real time. Using treating therapists to count motions would be expected to degrade clinical remediation, and using independent observers would be expected to outstrip financial and personnel resources at most institutions. Even with offline video analysis, the identification and counting of functional motions is prohibitively time- and personnel-intensive[23]. The impracticality of manual tallying thus limits widespread adoption in research or clinical settings, and therefore curtails the identification and implementation of effective dosing parameters.

To overcome these limitations, we present the Primitive Sequencing pipeline (PrimSeq), a deep learning-based framework to automatically identify and count functional motion repetitions in rehabilitation training. Inspired by deep learning methods for speech recognition, our approach uses a sequence-to-sequence model to generate sequences of functional primitives[24]. Functional primitives are fundamental motions that can be strung together to constitute a functional activity, akin to words in a paragraph[23]. We previously found that rehabilitation activities can be broken down into five main classes of primitives: *reach*, *reposition*, *transport, stabilize*, *idle*[23, 25, 26]. Here, we sought to classify and count these units of motion with our pipeline.

PrimSeq encompasses three main steps: (1) capture of upper body motion during rehabilitation with wearable inertial measurement units (IMUs), (2) generation of primitive sequences from IMU data with the trained deep learning model, and (3) tallying of primitives with a counting algorithm. We applied PrimSeq to chronic stroke patients performing a battery of rehabilitation activities. Our deep learning-based approach robustly identified primitives in various rehabilitation activities and in patients with a range of motor impairment, and outperformed other activity-recognition methods. Our IMU-based motion capture was also well tolerated by patients, and PrimSeq considerably diminished the time and burden of quantifying functional repetitions.

**Results**

To collect a variety of functional primitives to train the deep learning model, we recorded and labeled functional motion from 41 chronic stroke patients with UE paresis (Fig. 1; Supplementary Table 1). While patients performed common rehabilitation activities (Fig. 1A; Supplementary Table 2), we captured upper-body motion with an array of wearable inertial measurement units (IMUs) and video cameras (Fig. 1B). To generate ground truth labels, trained human coders viewed the videotaped activities and subdivided them into functional primitives (Fig. 1C). The coders annotated the beginning and end of each primitive on the video, which applied a primitive label to the corresponding segment of IMU data. Interrater reliability of primitive labeling was high between the coders and expert (Cohen's $K \geq 0.96$). We split the labeled IMU data into a training set (n = 33 patients; 43,429 primitives) and an independent test set (n = 8 patients; 11,502 primitives), balancing for impairment level and paretic side (Supplementary Table 1; data are available here https://simtk.org/projects/primseq). We note that the number and variety of primitives, not the number of subjects, makes this dataset robust for development of deep learning approaches.

We used the labeled training set for model fitting, parameter adjustment, and hyperparameter tuning. We designed a sequence-to-sequence (Seq2Seq) deep learning model that predicts primitives from IMU data patterns. Seq2Seq consists of two recurrent neural networks (RNNs) that map a window of motion data to a sequence of primitives (see Methods for architecture design and training details).

The trained Seq2Seq model is central to the PrimSeq pipeline, which has three main steps (Fig. 2). First, IMU data from rehabilitation activities are recorded and divided into 6-second windows (Fig. 2A). Second, this windowed data is fed to the Seq2Seq model, where an encoder RNN generates a single feature vector capturing relevant motion information, which is processed by the decoder RNN to estimate a sequence of primitives (Fig. 2B). Third, a counting algorithm removes any duplicates at window boundaries and tallies primitives from the sequences (Fig. 2C).

We used the previously unseen test set for an unbiased evaluation of the PrimSeq pipeline, employing ground truth labels to assess its primitive counting and classification performance.

**PrimSeq has high counting performance across primitive classes and activities**

We first examined the counting performance of PrimSeq in the rehabilitation activities (Fig. 3). In the separate primitive classes, the approach counted on average 282 reaches, 206 repositions, 408 transports, 291 stabilizations, and 258 idles (Fig. 3A). In the separate rehabilitation activities, the approach counted on average 40–308 primitives overall (Fig. 3B). To assess the similarity of predictions to the actual number of primitives performed, we compared predicted versus ground truth counts per primitive class and activity. The approach generated primitive counts that were 86.1–99.6% of true counts for the separate primitive classes (Fig. 3A) and 79.1–109.1% of true counts for the separate activities (Fig. 3B). We also examined counting errors at the single-subject level, finding that they were consistently low for primitive classes (reach 7.2 ± 9.9%, reposition 13.3 ± 9.2%, transport 0.4 ± 17.7%, stabilization 8.0 ± 42.7%, and idle 6.6 ± 10.5%) and for activities (shelf task -10.7 ± 26.0%, tabletop task -2.3 ± 13.4%, feeding 5.9 ± 16.7%, drinking 5.5 ± 25.0%, combing 11.8 ± 7.8%, donning glasses 14.5 ± 23.7%, applying deodorant 11.5 ± 26.2%, face washing 12.2 ± 30.1%, and tooth-brushing 6.3 ± 31.2%). In most cases, the variance in predicted counts could be attributed to inter-individual variance in true counts. However, for stabilizations, an excess variance in predicted counts could be explained by increased prediction errors, which we further discuss below.

**Seq2Seq error examination**

PrimSeq generated primitive counts that closely approximate true counts, but gross tallies do not reveal if the Seq2Seq model identified primitives when they were actually performed. For example, the model may fail to predict a reach that happened at a given time but may predict a reach that did not happen moments later; the net result of these two errors is that a reach is spuriously credited to the count. We thus examined the nature of predictions made by Seq2Seq (Fig. 4). We compared the predicted and ground truth sequences using the Levenshtein algorithm[27], identifying two types of prediction errors: false negatives and false positives (Fig. 4A). A false negative occurred when Seq2Seq did not predict a primitive that actually happened because it erroneously missed the primitive (deletion error) or erroneously predicted another primitive class (swap-out error). A false positive occurred when Seq2Seq predicted a primitive that did not actually happen because it erroneously added the primitive (insertion error) or erroneously predicted this primitive class (swap-in error).

We examined the frequency of these error types with respect to true counts in each primitive class, which adjusts for differences in number of primitives performed (Fig. 4B). Seq2Seq had a modest frequency of deletion errors for most primitives (10.6–13.6%) except stabilizations (25.5%) and a modest frequency of swap-out errors (5.9–11.2%) for all primitives. Seq2Seq also had a modest frequency of insertion errors (4.6–13.3%) and swap-in errors (4.0–13.4%) for all primitives.

Overall, Seq2Seq had a tolerable error rate, but had the most difficulty classifying stabilizations. The motion phenotype of stabilizations[23] could lead to different classification errors. Seq2Seq could blend the minimal motion of stabilizations into the beginning or end of an adjacent motion-based primitive (i.e., reach, reposition, transport), leading to its deletion. Conversely, the model could mistake periods of diminished motion in adjacent primitives as a stabilization, leading to an insertion or swapping-in. In addition, the model had an increased frequency of swapping-in stabilizations for idles (Supplementary Fig. 1). This error could be attributed to the lack of IMU finger data necessary to identify grasp, which is a major phenotypic distinction between these two minimal-motion primitives[23].

**Seq2Seq classification performance**

To assess the overall classification performance of Seq2Seq to predict primitives, we computed sensitivity and false discovery rate (FDR). Sensitivity represents the proportion of true primitives that were correctly predicted and included in the count. The FDR, a measure of overcounting, represents the proportion of predicted primitives that were incorrectly predicted and included in the count. We assessed Seq2Seq classification performance for separate primitive classes, activities, and patient impairment levels (Fig. 5).

**Seq2Seq classifies most primitives well**

We first examined Seq2Seq classification performance for each primitive class (Fig. 5A). The model had high mean sensitivities for most primitives (0.76–0.81) except stabilizations (0.64). The model also had low mean FDRs for most primitives (0.11–0.16) except stabilizations (0.23). For stabilizations, distinct prediction errors drove the modest classification performance: their spurious removal (false negatives) decreased sensitivity, whereas their spurious addition (false positives) increased overcounting.

**Seq2Seq classifies primitives best in structured functional activities**

We next examined Seq2Seq classification performance for each activity (Fig. 5B). The model had excellent performance with structured activities, such as moving an object to fixed locations on a shelf (mean sensitivity 0.92, FDR 0.07) and tabletop (mean sensitivity 0.91, FDR 0.06). Its performance declined with more naturalistic activities, such as drinking (mean sensitivity 0.74, FDR 0.16) and feeding (mean sensitivity 0.74, FDR 0.19). Seq2Seq had its lowest performance in the tooth-brushing activity (mean sensitivity 0.62, FDR 0.21).

**Seq2Seq performs well for patients with mild to moderate UE impairment**

We also examined if Seq2Seq performance was affected by impairment level (Fig. 5C). Seq2Seq had a stable sensitivity (0.71–0.82) that did not vary with UE-FMA score (Spearman's correlation ($\rho(6)$) = 0.54, p = 0.171, 95% confidence interval (CI) [-0.12, 0.87]). The FDR ranged more widely (0.31–0.64) and showed a trend for decreasing as UE-FMA scores increased ($\rho(6)$ = -0.69, p = 0.069, 95% CI [-0.11, -0.91]). This trend was driven by one patient (UE-FMA 37), whose stabilizations and transports were excessively overcounted by Seq2Seq.

**Seq2Seq outperforms benchmarks**

Finally, we benchmarked Seq2Seq against competitive models used in human action recognition: convolutional neural network (CNN), action segment refinement framework (ASRF), and random forest (RF, Fig. 6)[25, 28, 29]. The CNN and RF made predictions at each 10-ms time point, which were smoothed to generate primitive sequences[30]. ASRF is a state-of-the-art action recognition method that directly generates primitive sequences. We aggregated patients, primitive classes, and activities to examine the overall sensitivity and FDR of each model. We also examined the $F_1$ score, the harmonic mean between sensitivity and precision (1-FDR), which reflects global classification performance.

Seq2Seq had the highest sensitivity of the models tested (Seq2Seq, 0.77; benchmark models, 0.50–0.73). Seq2Seq also had a lower FDR than CNN and RF, but was marginally outperformed by ASRF (Seq2Seq, 0.17; benchmark models, 0.16–0.21). Finally, Seq2Seq had the highest $F_1$ score, indicating the best overall classification performance of the models tested (Seq2Seq, 0.799; benchmark models, 0.61–0.76).

**PrimSeq provides a practical solution to count functional motion repetitions**

Finally, we examined the practicality of using the PrimSeq pipeline. From the motion capture standpoint, most patients reported minimal difficulty with donning and calibrating the IMU array and minimal discomfort with wearing them (median scores 0 and 1, respectively, on an 11-point visual analog scale; Supplementary Fig. 2). IMU setup time was minimal, requiring on average 12.9 ± 4.1 minutes. Electromagnetic sensor drift was also late to emerge during recording, occurring after 63.7 ± 28.1 minutes, with recalibration taking 1–2 minutes. Collectively, these observations indicate that an IMU system could support motion capture that is well tolerated, efficient, and stable.

From the primitive identification standpoint, the trained Seq2Seq model and counting algorithm identified and tallied primitives 366 times faster than human coders. For example, to process 6.4 hours of recorded activities, Seq2Seq had an execution time of 1.4 hours (13 s per minute of recoding) whereas trained human coders required 513.6 hours or 12.8 workweeks (4,815 s per minute of recording). Furthermore, if inference is made on a high-performing

abstract

computer with an advanced graphical processing unit, processing lags between incoming IMU data and primitive prediction are estimated at ~15 seconds. These observations indicate that PrimSeq could provide near-immediate feedback about primitive counts, which would help patients and therapists meet training goals during rehabilitation sessions[31].

**Discussion**

To date, the measurement of training repetitions in UE stroke rehabilitation has been an elusive technical challenge, hampered by imprecision or impracticality. To address these obstacles, we developed PrimSeq, an approach that integrates wearable IMU data, a trained deep learning model, and a counting algorithm to quantify functional primitives performed during rehabilitation. We found that PrimSeq generated accurate primitive counts across primitive classes and activities. We also found that the Seq2Seq model also had a moderate-to-high classification performance across primitive classes, rehabilitation activities, and impairment levels, outperforming state-of-the-art action recognition models. Finally, we found that PrimSeq was a practical approach for UE motion capture and processing. These results indicate that PrimSeq provides a pragmatic solution for the accurate identification of motion content and quantity during rehabilitation.

Our approach has some limitations to consider. PrimSeq identifies functional motions but does not measure how normally they are performed. This information is important for tracking recovery and tailoring rehabilitation. Future work could characterize and reference the normative kinematics of the primitive classes, which could generate continuous measurements of abnormal primitive performance. In addition, PrimSeq was trained on motion from chronic stroke patients performing a circumscribed battery of rehabilitation activities. To increase clinical utility, additional model training and refinement could be undertaken "in the wild" with subacute stroke patients undergoing inpatient rehabilitation. An expanded variety of sampled primitives could be expected to boost classification performance on unstructured activities, and make PrimSeq robust for application in different recovery stages. Finally, the classification performance of Seq2Seq was limited in some cases (e.g. stabilizations, tooth-brushing activity). Future work could employ alternative deep learning models with explainable artificial intelligence to identify sources of confusion, which could then be targeted to improve classification performance.

In conclusion, we present a novel pipeline that measures functional motion repetitions in UE rehabilitation activities. PrimSeq is a foundational step toward the precise and pragmatic quantification of rehabilitation dose, and overcomes considerable time, personnel, and financial barriers. Our approach has the potential to support the rigorous rehabilitation research and clinical delivery that are vitally needed to improve stroke outcomes.

## Methods

### Subjects

We studied 41 chronic stroke patients with upper extremity (UE) paresis. Patients gave written informed consent to participate. This study was approved by the Institutional Review Board at New York University Langone Health, in accordance with the Declaration of Helsinki. Patient demographics and clinical characteristics are reported in Supplementary Table 1.

### Enrollment criteria

Eligibility was determined by electronic medical records, patient self-report, and physical examination. Patients were included if they were ≥ 18 years old, premorbidly right-handed, able to give informed consent, and had unilateral motor stroke with contralateral UE weakness scoring < 5/5 in any major muscle group[32]. Patients were excluded if they had: hemorrhagic stroke with mass effect, or subarachnoid or intraventricular hemorrhage; traumatic brain injury; musculoskeletal, major medical, or non-stroke neurological condition that interferes motor function; contracture at shoulder, elbow, or wrist; moderate UE dysmetria or truncal ataxia; apraxia; visuospatial neglect; global inattention; or legal blindness. Stroke was confirmed by radiographic report. Lesions in non-motor areas or the opposite hemisphere were allowed barring bilateral weakness. Both ischemic and hemorrhagic stroke were included, as motor deficits do not substantially differ between the two types[33]. Stroke patients were chronic (> 6 months post-stroke) except for two patients (3.1 and 4.6 months post-stroke).

**Primitive dataset generation**

Patients participated in two to three sessions lasting ~2.5 hours. Sessions were typically one to three days apart (average 2.6 days). At the first session, we recorded patient height and measured UE impairment level with the UE Fugl-Meyer Assessment (FMA), where a higher score (maximum 66) indicates less impairment[34]. Patients then performed five trials of nine rehabilitation activities while their upper body motion was recorded (Supplementary Table 2; Fig. 1A). We identified activities using a standardized manual of occupational therapy (OT)[35]. From these, we identified activities commonly practiced during inpatient stroke rehabilitation through survey of seven OTs with expertise in stroke rehabilitation. Patients were seated in front of a workspace (table or sink counter) at a distance that allowed the nonparetic UE to reach, without trunk flexion, to the furthest target object. Workspaces were adjusted to standard heights (table, 76 cm; counter, 91 cm). We placed the target objects at fixed locations using marked, laminated cardboard mats (table) or measured distances (counter). We used standardized instructions that outlined the major goals of the activity. Because most activities in the battery are bimanual, we instructed patients to use their paretic UE to the best of their ability.

**Motion capture**

To record patient motion, we affixed nine inertial measurement units (IMUs; Noraxon, USA) to the C7 and T10 spine, pelvis, and both hands using Tegaderm tape (3M, USA) and to both arms and forearms using Velcro straps (Fig. 1A). Each IMU is small (length: 37.6 mm; width: 52.0 mm; height: 18.1 mm), lightweight (34 g), and captures 3D linear acceleration, 3D angular velocity, and 3D magnetic heading at 100 Hz. The motion capture software (myomotion, Noraxon, USA) applies a Kalman filter to the angular velocities to generate 3D unit quaternions for each sensor[36]. We used coordinate transformation matrices to transform the generated quaternions to a sensor-centric framework, which represents the rotation of each sensor around its own axes. The motion capture software also applies a proprietary height-scaled skeletal model to the IMU data to generate 22 anatomical angles of the upper body (Supplementary Table 3). The motion capture system thus generates a 77-dimensional dataset every 10 ms, displayed on a software interface alongside an avatar of the patient (Fig. 1B).

The motion capture system records patient motion with high precision (accelerometry accuracy ± 0.001 g; gyroscopic accuracy ± 1.25°; anatomical angle accuracy ± 2°), performing as well as the gold-standard optical system[37]. We monitored online for electromagnetic sensor drift by visually inspecting the joint angle data for baseline shifts and ensuring that avatar motions matched those of the patient. Patients were immediately recalibrated if drift was observed. Recalibration required standing with UEs straight at the sides (arms, forearms, and wrists in neutral position with elbows extended) and took less than two minutes.

We recorded UE motion with two high-speed (60 Hz), high-definition (1088 × 704 resolution) cameras (Ninox, Noraxon) positioned orthogonally less than two meters from the patient (Fig. 1A). The cameras have a focal length of f4.0 mm and a large viewing window (length: 2.5 m; width: 2.5 m; Fig. 1B). The cameras ran on the same clock as the IMUs and video and IMU recordings were synchronized.

**Data Labeling**

Human coders identified the functional primitives performed in the rehabilitation activities. The five classes of functional primitives are *reach* (UE motion into contact with a target object), *reposition* (UE motion proximate to a target object), *transport* (UE motion to convey a target object), *stabilize* (minimal UE motion to keep a target object still), and *idle* (minimal UE motion to stand at the ready near a target object). Coders were trained on a functional motion taxonomy that operationalizes primitive identification[23]. The coders used the video recordings to identify and label the start and end of each primitive, which simultaneously segmented and labeled the synchronously recorded IMU data. To ensure the reliable labeling of primitives, an expert (AP) inspected one-third of all coded videos. Interrater reliability between the coders and expert was high, with Cohen's K coefficients ≥ 0.96. Coders took on average 79.8 minutes to annotate one minute of recording.

We split the resulting ground truth dataset of into a training set (n = 33 patients; 43,429 primitives) and test set (n = 8 patients; 11,502 primitives) to independently train and test the deep learning model. Patient selection was random but constrained to balance impairment level and paretic side. The IMU dataset, including its data splits, are available on https://simtk.org/projects/primseq.

**Deep learning model development**

Inspired by speech recognition models[38], we used a sequence-to-sequence (Seq2Seq) deep learning model to perform the task of predicting primitive sequences. To handle the higher dimensionality of the IMU data, we increased the model's input nodes to 77 (from 40 for speech) and increased the hidden dimensionality to 3072 (from 512 for speech). To provide sufficient context of the time series given lower sampling rates of 100 Hz for motion data (versus 16,000 Hz for speech), we expanded the window size of the input data to 6 s (from 10 ms for speech).

Seq2Seq performs primitive identification in two steps (Fig. 2). The model first encodes the data window using a bidirectional encoder recurrent neural network (RNN) to generate a 3072-element feature vector. The encoder RNN reduces the dimensionality of the high-dimensional IMU data, which enables it to learn relevant features from the IMU data for the downstream task of sequence prediction. The model then decodes this feature vector using a unidirectional decoder RNN to generate the sequence of primitives. The generated sequence of primitives is then passed through a counting algorithm that counts the functional motion repetitions while removing duplicate primitives at the window boundaries. Additional model details are presented in recent work[24].

We trained Seq2Seq by minimizing a loss function based on the cross-entropy between the predicted and ground truth primitive sequences using the Adam optimizer[39]. We used a learning rate of $5 \times 10^{-4}$. Because primitive overcounting may lead to accidental under-training of patients in rehabilitation, we prioritized keeping the average false discovery rate (FDR) < 20% while maximizing the average sensitivity during model training. We ensured this balance by stopping the model training early based on the Action Error Rate (AER), computed as the total number of changes needed on the predicted sequence to match the ground truth, normalized to the length of the ground truth sequence.

We used a window size of 6 s for primitive prediction with Seq2Seq. During model training, the middle 4 s of the window was predicted, with the flanking 1 s of data providing the model additional temporal context for prediction. We further maximized the training data by adding a window slide of 0.5 s, which also helped the model learn primitive boundaries. During model testing, the window size was 6 s and middle 4 s of the window was predicted. To enable the flanking during model testing, we set the slide to 4 s. A preliminary experiment was performed with window sizes of 2 s, 4 s, and 8 s. The window size of 6 s resulted in the lowest validation AER.

We selected and cross-validated the hyperparameters for Seq2Seq with four different splits of the training set. In each split, 24 or 25 patients were used for training and 9 or 8 patients were used for validation. We selected the hyperparameters for each of the four models based on their validation AERs. Each split yields a separate model that generates independent prediction probabilities per primitive. The prediction probabilities from the four models were averaged, or ensembled, and the primitive with the highest probability was taken as the Seq2Seq prediction. The ensembled prediction was also fed back into the four models to inform the next prediction in the data window.

After Seq2Seq training and hyperparameter estimations were done on the training set, we applied the trained Seq2Seq model to the test set to assess its counting and classification performance in data from previously unseen patients. The test set was not used for feature selection, preprocessing steps, or parameter tuning. Code implementing the model, including instructions for training and hyperparameter selection, and comparisons with other action-recognition methods on benchmark datasets are available on [https://github.com/aakashrkaku/seq2seq_hrar](https://github.com/aakashrkaku/seq2seq_hrar) . The model predictions with respect to ground truth primitives can be visualized with this video of patient motion (Supplementary Video).

**Analysis of counting performance**

To assess the counting performance of PrimSeq, we quantified the predicted and actual counts of primitives and normalized predicted to ground truth counts for each patient in the test set. To assess counting performance per primitive class, we combined counts from all activities. To assess counting performance per activity, we combined counts from all primitive classes. Counting performance is presented as the mean percent and standard deviation of true counts for the test set. We also examined counting error at the single-subject level, because mean values may obscure erroneous counting (e.g. an average of under-counts and over-counts would wash out errors). We calculated these counting errors as the difference between true and predicted counts per subject, normalized to true count. We report these single-subject counting errors as mean percent and standard deviation for the test set.

**Analysis of prediction outcomes and error frequency**

To examine the nature of predictions that Seq2Seq made per primitive class, we combined activities and compared predicted against ground truth sequences for each patient in the test set. We used the Levenshtein sequence-comparison algorithm to match the predicted and ground truth primitive sequences[27]. This step generated the prediction outcomes of false negative (FN), false positive (FP), or true positive (TP).

False negatives, or primitives spuriously removed from the prediction, could arise from a deletion error (the model did not predict a primitive that actually happened) or a swap-out error (the model did not predict the actual primitive but instead predicted an incorrect class). False positives, or primitives spuriously added to the prediction, could arise from an insertion error (the model predicted a primitive that did not actually happen) or a swap-in error (the model predicted an incorrect primitive class instead of the actual primitive). True positives were primitives that were correctly predicted.

We examined the frequency and type of prediction errors (FN, FP) made by Seq2Seq, normalizing prediction errors to ground truth counts to adjust for different quantities of primitives. The frequencies of prediction errors are presented as mean and standard deviation across test set patients. To further assess which classes of primitives were mistaken for each other, we generated a confusion matrix to examine swap-out and swap-in errors (Supplementary Fig. 1).

**Analysis of classification performance**

To examine the classification performance of Seq2Seq, we computed the classification performance metrics of sensitivity and false discovery rate (FDR) with respect to primitive class, activity, and impairment level.

Sensitivity, also known as true positive rate or recall, represents the proportion of ground truth primitives that were correctly predicted. It is calculated as:

$$Sensitivity = \frac{TP}{TP + FN}$$

FDR, a type of overcount, represents the proportion of predicted primitives that were incorrectly predicted. It is calculated as:

$$FDR = \frac{FP}{TP + FP}$$

To calculate sensitivity and FDR per primitive class, we combined prediction outcomes (i.e., TP, FN, and FP) from all activities for each test set patient (Fig. 5A). To calculate sensitivity and FDR per activity, we combined prediction outcomes from all primitives for each test set patient (Fig. 5B). Sensitivity and FDR are reported as means and standard deviations across test set patients.

Finally, to assess classification performance at different levels of UE impairment, we combined prediction outcomes from all primitives and activities for each test set patient and calculated sensitivity and FDR (Fig. 5C). We examined if these performance metrics varied with ordinal UE FMA scores using Spearman's correlation ($\rho$).

**Model benchmarking**

We compared Seq2Seq against three benchmark models used in human action recognition: convolutional neural network (CNN), random forest (RF), and action segment refinement framework (ASRF).

We examined the classification performance of a CNN that we previously developed to predict primitives from IMU data[25]. Each layer in the CNN computes linear combinations of outputs of the previous layer, weighted by the coefficients of convolutional filters. The model includes an initial module that helps to map different physical quantities captured by IMU system (e.g., accelerations, joint angles, and quaternions) to a common representation space. The model also uses adaptive feature-normalization to increase the robustness of the model to shifts in the distribution of the data, which can occur when the model is applied to new patients.

We also examined the classification performance of RF, a conventional machine learning model, which has previously been used for human activity recognition[28], including distinguishing functional from nonfunctional motion using wrist-worn sensors[16-18]. RF uses a number of decision trees on randomly selected sub-samples of the dataset to make predictions. We input into the model a set of statistical features for each data dimension, including its mean, maximum, minimum, standard deviation, and root mean square. These features capture useful information for motion identification, such as the energy and variance of the motion.

CNN and RF generate primitive predictions at each 10-ms time point. To generate primitive sequences, we smoothed the pointwise predictions of these models using a weighted running average approach. To perform the smoothing, we used a Kaiser window[30] whose parameters (window size and relative sidelobe attenuation) were selected using the best validation performance.

We also examined the classification performance of ASRF, a state-of-the-art deep learning model for action recognition[29]. ASRF is composed of two CNN modules: a segmentation module and a boundary detection module. The segmentation module performs the pointwise predictions of the primitives and the boundary detection module detects the boundaries of the primitives. These pointwise primitive predictions are combined with the detected boundaries for smoothing and final sequence generation. During smoothing, the model takes the most frequent pointwise prediction between two detected boundaries as the final prediction for that segment.

To benchmark Seq2Seq against these alternative models, we compared their overall sensitivity, FDR, and the $F_1$ score. We combined prediction outcomes from all patients, primitives, and activities. The $F_1$ score, a balance between sensitivity and FDR, captures the global classification performance of a model. The $F_1$ score ranges between 0 and 1, and a value of 1 indicates perfect classification. It is calculated as:

$$F_1 = 2 \left( \frac{sensitivity\ (1 - FDR)}{sensitivity + (1 - FDR)} \right) = 2 \left( \frac{TP}{TP + 0.5(FN + FP)} \right)$$

**Practicality assessment of PrimSeq**

To assess the practicality of PrimSeq, we first examined whether patients found the IMUs challenging to wear. We used a visual analogue scale (VAS) ranging from 0 (least) to 10 (most) to examine if donning IMUs (application and calibration) was difficult or wearing IMUs was uncomfortable. VAS scores were obtained at the end of each session and are reported as median and range across patients (Supplementary Fig. 2). We recorded the time to don and calibrate the IMUs in a subset of 10 patients, reported as mean and standard deviation. We also recorded the onset time of electromagnetic sensor drift in all patients, reported as mean and standard deviation, and the time needed for recalibration, reported as a range.

Finally, to compare the labeling speed of trained human coders against the trained Seq2Seq model, we recorded how long humans and the model took to label all activities from a subset of 10 patients (6.4 h of recordings). Seq2Seq processed the IMU data on a high-performing computer with an advanced graphical processing unit (GPU, 10 trillion floating-point operations per second, memory bandwidth of 900 GB/s). Total processing times are reported. We estimated processing lags between incoming IMU data and model predictions by summing the time of each interstitial operation: transferring data between the IMUs and myoMotion receiver, calling the API, preprocessing the data (quaternion transformation, z-score normalization), sizing the data windows, predicting the primitives on the GPU, and displaying and storing the data.

**Acknowledgements**

This work was funded by the American Heart Association/Amazon Web Service postdoctoral fellowship 19AMTG35210398 (A.P.), NIH R01 LM013316 (C.F.G., H.S.), NIH K02 NS104207 (H.S.), NIH NCATS UL1TR001445 (H.S.), and NSF NRT-HDR 1922658 (A.K., C.F.G.). We wish to thank Emily Fokas for assistance with preparing figures, Huizhi Li for software development to visualize model predictions, and the following for their assistance with annotating videos: Ronak Trivedi, Sanya Rastogi, Adisa Velovic, Vivian Zhang, Candace Cameron, Nicole Rezak, Sindhu Avuthu, Chris Yoon, Sirajul Islam, Caprianna Pappalardo, Alexandra Alvarez, Bria Barstch, Tiffany Rivera, and Courtney Nilson. We thank Drs. Jose Torres and Cen Zhang for assistance with identifying stroke patients.

## Author contributions

H.S. and C.F.G. conceived the project. A.P. and A.W. built the motion capture setup and created the rehabilitation battery with guidance from D.N. and H.S. A.P., A.V., N.P., and A.W. collected patient data and annotated videos. A.P., A.V., and N.P. quality-checked video annotations. A.K. and C.F.G. conceived the sequence-to-sequence architecture. A.K., H.R., and K.V. developed the model architecture and evaluated its performance with guidance from C.F.G. A.P. and A.K. performed data analysis. A.P., A.K., C.F.G. and H.S. led the project, and A.P. and H.S. wrote the paper with input from all authors.

## Conflict of interest

The authors have no conflicts of interest to declare.

This manuscript has not been peer-reviewed.19. Birkenmeier, R.L., Prager, E.M. & Lang, C.E. Translating animal doses of task-specific training to people with chronic stroke in 1-hour therapy sessions: a proof-of-concept study. *Neurorehabilitation Neural Repair* **24**, 620-635 (2010).
20. Lang, C.E., MacDonald, J.R. & Gnip, C. Counting Repetitions: An Observational Study of Outpatient Therapy for People with Hemiparesis Post-Stroke. *J. Neurol. Phys. Ther.* **31**, 3-10 (2007).
21. Lang, C.E. et al. Dose response of task-specific upper limb training in people at least 6 months poststroke: A phase II, single-blind, randomized, controlled trial. *Ann. Neurol.* **80**, 342-354 (2016).
22. Waddell, K.J., Birkenmeier, R.L., Moore, J.L., Hornby, T.G. & Lang, C.E. Feasibility of high-repetition, task-specific training for individuals with upper-extremity paresis. *Am. J. Occup. Ther.* **68**, 444-453 (2014).
23. Schambra, H.M. et al. A Taxonomy of Functional Upper Extremity Motion. *Frontiers in Neurology - Neurorehabilitation* **10**, 857 (2019).
24. Kaku, A. et al. Sequence-to-Sequence Modeling for Action Identification at High Temporal Resolution. *ArXiv* **abs/2111.02521** (2021).
25. Kaku, A. et al. Towards data-driven stroke rehabilitation via wearable sensors and deep learning. *Proceedings of machine learning research* **126**, 143-171 (2020).
26. Guerra, J. et al. Capture, learning, and classification of upper extremity movement primitives in healthy controls and stroke patients. *IEEE ... International Conference on Rehabilitation Robotics : [proceedings]* **2017**, 547-554 (2017).
27. Levenshtein, V.I. Binary codes capable of correcting deletions, insertions, and reversals. *Soviet physics doklady* **10**, 707-710 (1966).
28. Kwapisz, J.R., Weiss, G.M. & Moore, S.A. Activity recognition using cell phone accelerometers. *ACM SigKDD Explorations Newsletter* **12**, 74-82 (2011).
29. Ishikawa, Y., Kasai, S., Aoki, Y. & Kataoka, H. Alleviating Over-segmentation Errors by Detecting Action Boundaries. *2021 IEEE Winter Conference on Applications of Computer Vision (WACV)*, 2321-2330 (2021).
30. Kaiser, J. & Schafer, R. On the use of the I 0-sinh window for spectrum analysis. *IEEE Trans. Acoust.* **28**, 105-107 (1980).
31. Klassen, T.D. et al. Higher doses improve walking recovery during stroke inpatient rehabilitation. *Stroke* **51**, 2639-2648 (2020).
32. Medical Research Council of the United Kingdom Aids to Examination of the Peripheral Nervous System. (Pendragon House, Palo Alto, CA; 1978).
33. Saulle, M.F. & Schambra, H.M. Recovery and Rehabilitation after Intracerebral Hemorrhage. *Semin. Neurol.* **36**, 306-312 (2016).
34. Fugl-Meyer, A.R., Jaasko, L., Leyman, I., Olsson, S. & Steglind, S. The post-stroke hemiplegic patient. 1. a method for evaluation of physical performance. *Scandinavian Journal of Rehabilitation Medicine* **7**, 13-31 (1975).
35. Lang, C.E. & Birkenmeier, R.L. Upper-extremity task-specific training after stroke or disability: A manual for occupational therapy and physical therapy. (AOTA Press, 2014).
36. Sabatini, A.M. Quaternion-based extended Kalman filter for determining orientation by inertial and magnetic sensing. *IEEE Trans. Biomed. Eng.* **53**, 1346-1356 (2006).
37. Balasubramanian, S. in Center for Adaptive Neural Systems, Vol. Master's (Arizona State University, 2013).
38. Chan, W., Jaitly, N., Le, Q.V. & Vinyals, O. in arXiv preprint, Vol. arXiv:03167 (2015).
39. Kingma, D.P. & Ba, J. Adam: A method for stochastic optimization. *arXiv:1412.6980* (2014).
23

## a. Activity sampling

Motion capture setup

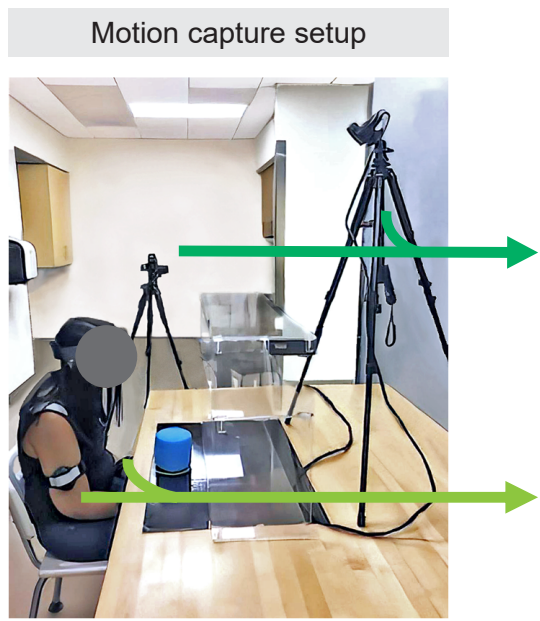

## b. Data recording

2-view video data

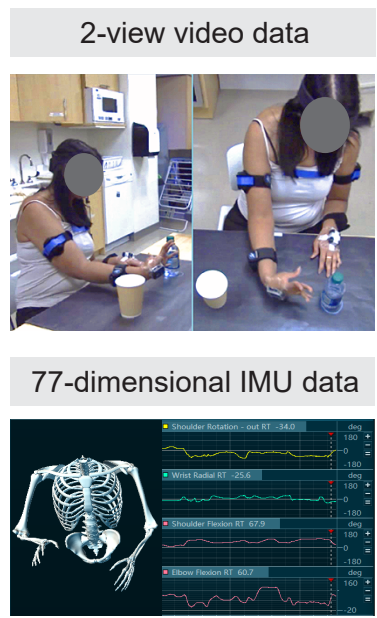

77-dimensional IMU data

## c. Primitive labeling

Video annotation

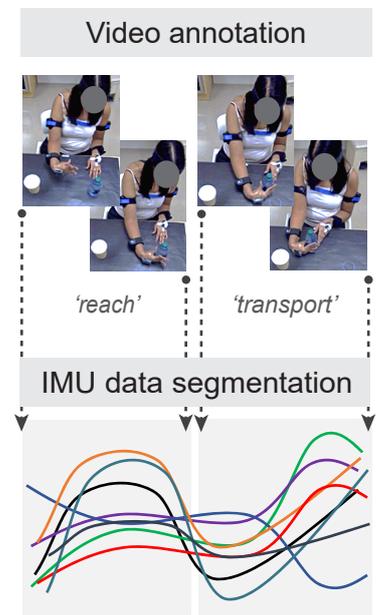

'reach'    'transport'

IMU data segmentation

**Figure 1. Functional motion capture and labeling. (A) Activity sampling.** As patients performed rehabilitation activities, functional motion was synchronously captured with two video cameras (dark green arrow) placed orthogonal to the workspace and nine inertial measurement units (IMUs, light green arrow) affixed to the upper body. **(B) Data recording.** The video cameras generated 2-view, high-resolution data. The IMU system generated 77-dimensional kinematic data (accelerations, quaternions, and joint angles). A skeletal avatar of patient motion and joint angle offsets were monitored for electromagnetic sensor drift. **(C) Primitive labeling.** Trained coders used the video recordings to identify and annotate functional primitives (dotted vertical lines). These annotations labeled and segmented the corresponding IMU data. Interrater reliability was high between the coders and expert (Cohen's K for reach, 0.96; reposition, 0.97; transport, 0.97; stabilization, 0.98; idle, 0.96).

### a. Motion capture

Activity performance | IMU data

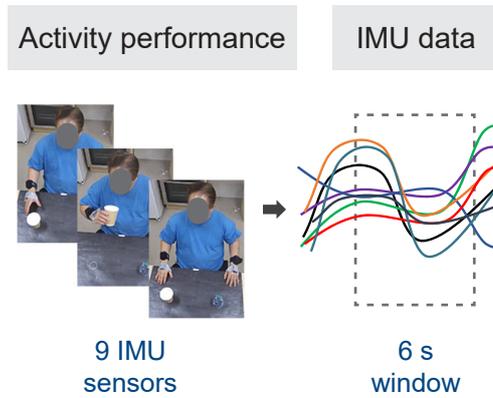

9 IMU sensors | 6 s window

### b. Primitive prediction

Sequence-to-sequence model

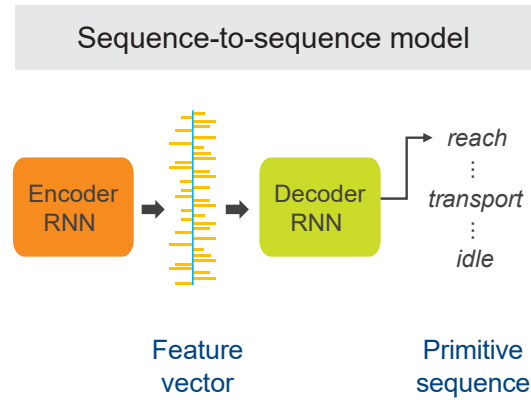

Feature vector | Primitive sequence

### c. Primitive count

Counting algorithm

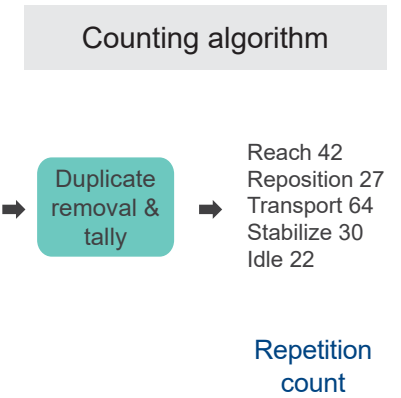

Repetition count

**Figure 2. The PrimSeq pipeline. (A) Motion capture.** Upper-body motion data are captured with IMUs during performance of rehabilitation activities. IMU data are divided into six-second windows. **(B) Primitive prediction.** The windowed IMU data are fed into the sequence-to sequence (Seq2Seq) deep learning model. Seq2Seq uses a recurrent neural network (RNN) to sequentially encode IMU data into a feature vector, which provides a condensed representation of relevant motion information. A second RNN then sequentially decodes the feature vector to generate the primitive sequence. **(C) Primitive count.** A counting algorithm then removes primitive duplicates at window boundaries and tallies the predicted primitives.

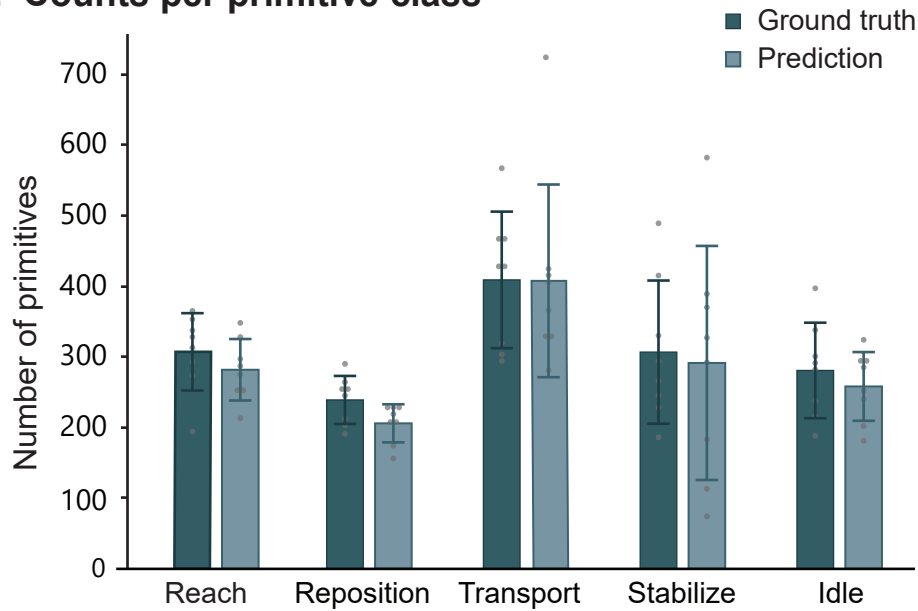

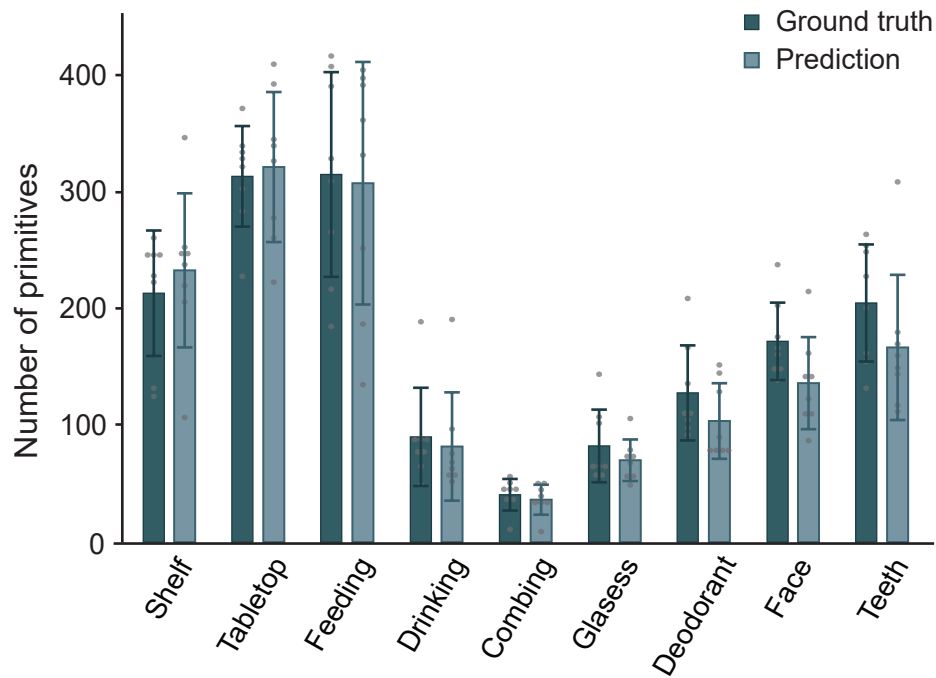

**Figure 3. Counting performance of PrimSeq.** Shown are the mean ± standard deviation of the ground truth and predicted counts for patients in the test set; each dot represents a single subject. **(A) Counts per primitive class.** Activities were combined. The pipeline generated counts that were similar to true counts for each primitive class (mean percent of true counts: reach, 92.2%; reposition, 86.6%; transport, 99.5%; stabilization, 91.1%; and idle, 93.3%). **(B) Counts per activity.** Primitive classes were combined. The pipeline generated counts that were similar to true counts for each activity (mean percent of true counts: shelf task, 109.1%; tabletop task, 102.5%; feeding, 97.6%; drinking, 93.7%; combing hair, 89.8%; donning glasses, 85.2%; applying deodorant, 81.1%; washing face, 79.1%; and brushing teeth, 81.4%).

## a. Sequence-matching

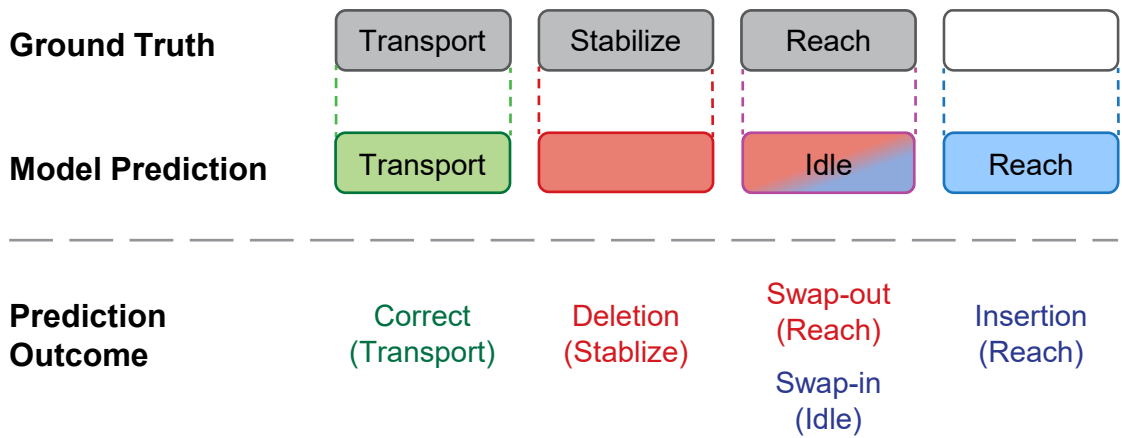

## b. Frequency of prediction errors per primitive class

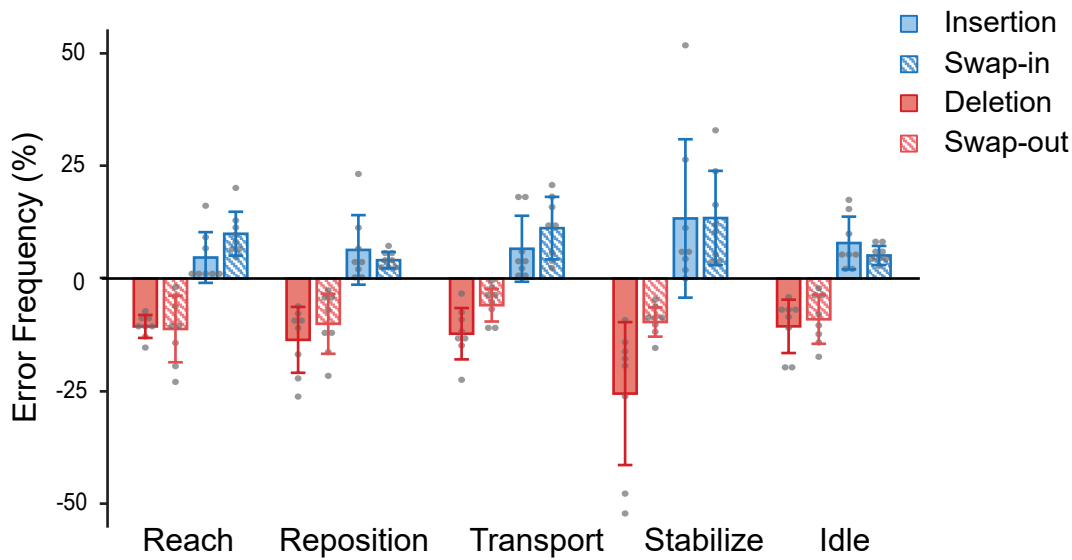

**Figure 4. Prediction errors by Seq2Seq. (A) Sequence matching.** Shown is a schematic depicting the different types of prediction outcomes, comparing the ground truth sequence (top row) against the predicted sequence (bottom row). Seq2Seq could produce a true positive (correct prediction; green), false negative (deletion or swap-out error; pink), or false positive (insertion or swap-in error; blue). In the example shown, transport was correctly predicted; stabilize was incorrectly deleted; reach was incorrectly swapped-out while idle was incorrectly swapped-in; and reach was incorrectly inserted. **(B) Frequency of prediction errors per primitive class.** Shown are the mean freqency ± standard deviation of prediction errors for patients in the test set; each dot represents a single subject. Activities were combined, and erroneous counts were normalized to ground truth counts in each primitive class. Deletion errors happened when primitives were incorrectly removed from the prediction, and occurred with modestly low frequency, except for stabilizations. Swap-out errors happened when primitives were incorrectly removed from the prediction and instead predicted as another class, and occurred with modestly low frequency. Insertion errors happened when primitives were incorrectly added to the prediction, and occurred with low frequency, except for stabilizations. Swap-in errors happened when primitives were incorrectly predicted instead of the actual primitive class, and occurred with modestly low frequency.

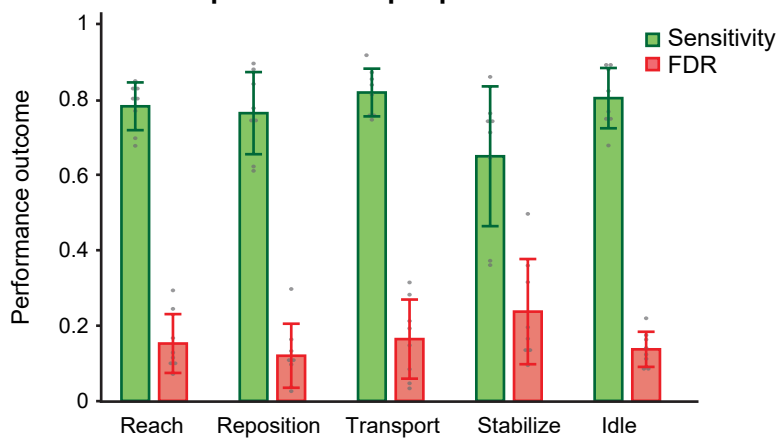
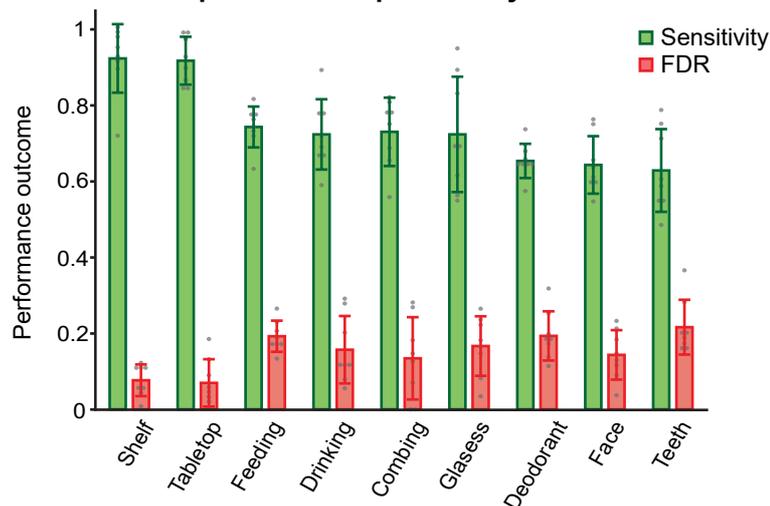
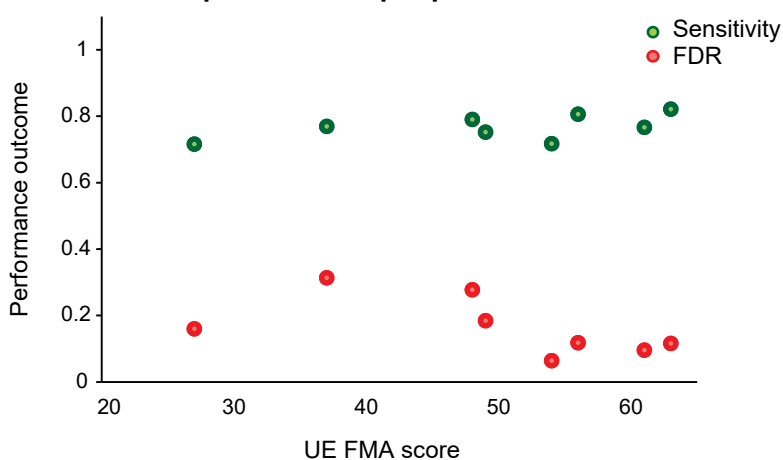

**Figure 5. Classification performance of Seq2Seq. (A) Classification performance per primitive class.** Prediction outcomes (FP, FN, TP) for activities were combined. Shown are mean ± standard deviation of Seq2Seq sensitivity and FDR for patients in the test set; each dot represents a single subject. Mean sensitivity was high and mean FDR was low for most primitives except stabilizations.
**(B) Classification performance per activity.** Prediction outcomes for primitive classes were combined. Shown are mean ± standard deviation of Seq2Seq sensitivity and FDR for patients in the test set; each dot represents a single subject. Mean sensitivity was high and mean FDR was low for structured activities such as the shelf and tabletop tasks, but were more modest for more complex activities. **(C) Classification performance per patient.** Prediction outcomes for activities and primitives were combined. Shown are sensitivity and FDR values per patient with respect to their upper extremity Fugl-Meyer Assessment (UE FMA) score. Seq2Seq sensitivity was not affected by impairment level (p = 0.171), but there was a trend for reduced FDRs with higher UE-FMA scores (p = 0.069), driven by one patient.

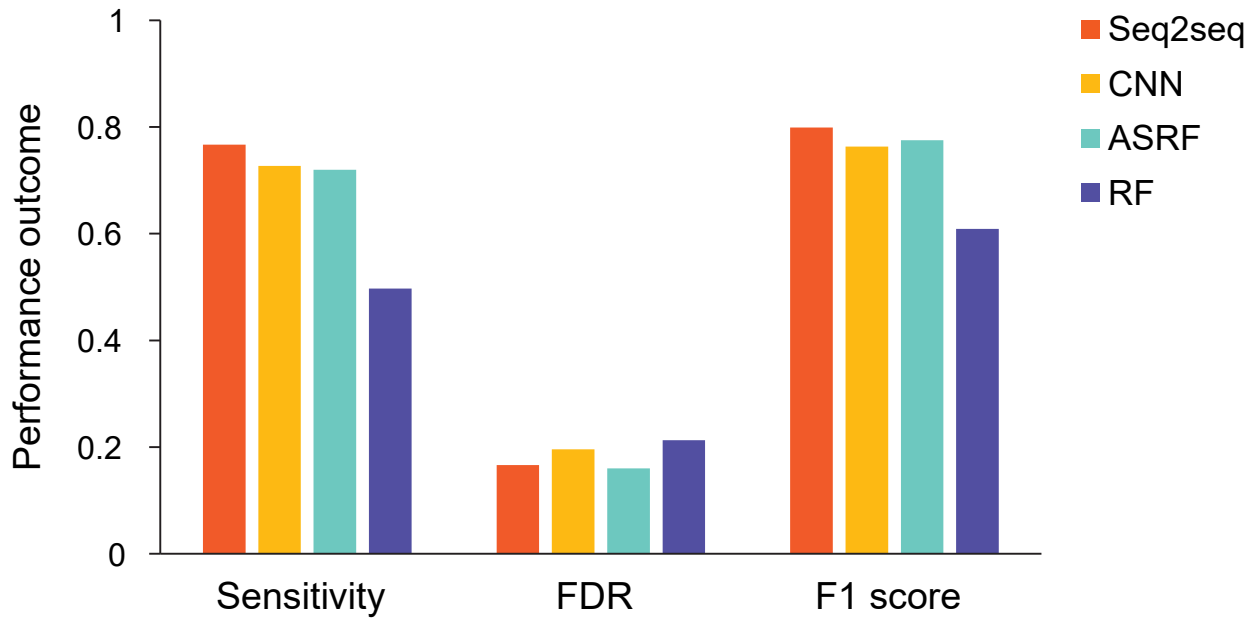

**Figure 6. Benchmarking Seq2Seq performance.** Prediction outcomes were aggregated from primitives, activities, and patients. Shown are the sensitivity, false discovery rate (FDR), and F1 score for Seq2Seq, convolutional neural network (CNN), action segment refinement framework (ASRF), and random forest (RF). Seq2Seq had the highest sensitivity (Seq2Seq, 0.767; CNN, 0.727; ASRF, 0.720; RF, 0.497). Seq2Seq FDR was lower than CNN and RF but marginally higher than ASRF (Seq2Seq, 0.166; CNN, 0.196; ASRF, 0.160; RF, 0.213). Seq2Seq had the highest F1 score (Seq2Seq, 0.799; CNN, 0.763; ASRF, 0.775; RF, 0.609).

| Patient characteristics | Training set | Test set |
|---|---|---|
| **Patient n** | 33 | 8 |
| **Primitive n** | 54,931 | 11,502 |
| **Age (Years)** | 56.3 (21.3-84.3) | 60.9 (42.6-84.3) |
| **Gender n** | 18 female: 15 male | 4 female: 4 male |
| **Stroke type** | 30 ischemic: 3 hemorrhagic | 8 ischemic |
| **Time since stroke (years)** | 6.5 (0.3-38.4) | 3.1 (0.4-5.7) |
| **Paretic side** | 18 left: 15 right | 4 left: 4 right |
| **UE FMA score (points)** | 48.1 (26-65) | 49.4 (27-63) |

**Supplementary Table 1. Demographics and clinical characteristics of patients.** The patient cohort (n = 41) was divided into a training set and test set, with no overlap. Impairment level was measured using the upper extremity Fugl-Meyer Assessment (UE FMA) score, with a maximum normal score of 66. Total n or average values with ranges are shown.

| Activity | Workspace | Target Object(s) | Instructions to Patient |
|---|---|---|---|
| **Shelf task** | On table, two Plexiglas shelves (length: 90 cm; width: 25.5 cm; height: 33, 53 cm) with three targets (22.5, 45, and 67.5 cm from left-most edge) placed 20 cm from table edge and center target (diameter: 5 cm) placed 2.5 cm from table edge | Toilet paper roll wrapped in self-adhesive wrap | Move the toilet paper back and forth between the center target and each target on the shelf, returning the arm to rest between each motion and the end. |
| **Tabletop task** | On table, horizontal circular array (diameter: 48.5 cm) of eight outer targets and center target (diameter: 5 cm) placed 2.5 cm from table edge | Toilet paper roll wrapped in self-adhesive wrap | Move the roll back and forth between the center and each outer target, returning the arm to rest between each motion and the end. |
| **Feeding** | On table, paper plate (diameter: 21.6 cm) placed at midline, 5 cm from table edge; utensils placed three cm from table edge and five cm from either side of plate; plastic sandwich baggie containing a slice of bread placed 25 cm from table edge, and 23 cm left of midline; and margarine packet placed 32 cm from table edge and 17 cm right of midline | Paper plate, fork, knife, re-sealable sandwich baggie, slice of bread, single-serve margarine packet | Pick up the sandwich baggie and open it, remove the bread and put it on the plate, open the margarine packet and spread margarine on the bread, cut the bread into four pieces, cut off and eat a small piece. |
| **Drinking** | On table, water bottle and paper cup placed 18 cm left and right of midline, respectively, and 25 cm from table edge | Water bottle (12 oz), cup (four oz) | Pick up the water battle and open it, pour some water into the cup, take a sip of water, place the cup on the table, and replace the cap on the bottle. |
| **Combing hair** | On table, comb placed at midline, 25 cm from table edge | Comb | Pick up the comb and comb both sides of the head. |
| **Donning glasses** | On table, pair of glasses placed at midline, 25 cm from table edge | Pair of glasses | Pick up the glasses, open the sides, put on the glasses, briefly place your hands down on the table, and then remove the glasses and place them on the table. |
| **Applying deodorant** | On table, deodorant placed at midline, 25 cm from table edge | Deodorant (solid, twist-base) | Pick up the deodorant, remove the cap, twist the base a few times, apply the deodorant to each armpit, replace the cap, untwist the base, place the deodorant on the table. |
| **Washing face** | At counter, small tub (length: 32.3 cm; width: 24.1 cm; depth: 2.5 cm) in sink; two folded washcloths on each side of counter next to sink, 30 cm from counter edge | Washcloths, tub, faucet handles | Fill the tub with water, take one washcloth and dip it into water, wring it, wipe each side of the face, and place it back on the counter. Then use the other washcloth to dry the face, and place it back on the counter. |
| **Brushing teeth** | At counter, toothpaste and toothbrush placed on either side of the sink, 30 cm from counter edge | Travel-sized toothpaste, toothbrush with built-up foam grip, faucet handles | Wet the toothbrush, open the toothpaste and apply it to the toothbrush, replace the cap on the toothpaste tube, brush teeth, rinse the toothbrush and mouth, and place the toothbrush back on the counter. |

**Supplementary Table 2. Activity battery**. These representative rehabilitation activities were used to generate an abundant sample of functional primitives for model training. Activity parameters include the workspace setup, target objects, and instructions to complete each task. The table and counter edges are their anterior edges closest to the patient. Patients could perform the actions within the activity in their preferred order.

| Joint | Anatomical angle |
|---|---|
| **Shoulder** | Shoulder flexion/extension<br>Shoulder internal/external rotation<br>Shoulder adduction/abduction<br>Shoulder total flexion[‡] |
| **Elbow** | Elbow flexion/extension |
| **Wrist** | Wrist flexion/extension<br>Forearm pronation/supination<br>Wrist radial/ulnar deviation |
| **Thorax** | Thoracic* flexion/extension<br>Thoracic* axial rotation<br>Thoracic* lateral flexion/extension |
| **Lumbar** | Lumbar[‡] flexion/extension<br>Lumbar[‡] axial rotation<br>Lumbar[‡] lateral flexion/extension |

**Supplementary Table 3: Anatomical upper body angles.** The motion capture system (myomotion, Noraxon, USA) used 9 IMUs and a proprietary height-scaled model to generate 22 upper body angles, shown in relation to their joint of *origin*. [‡]Shoulder total flexion is a combination of shoulder flexion/extension and shoulder ad-/abduction. *Thoracic angles are computed between the cervical (C7) and thoracic (T10) vertebrae. †Lumbar angles are computed between the thoracic vertebra and pelvis.

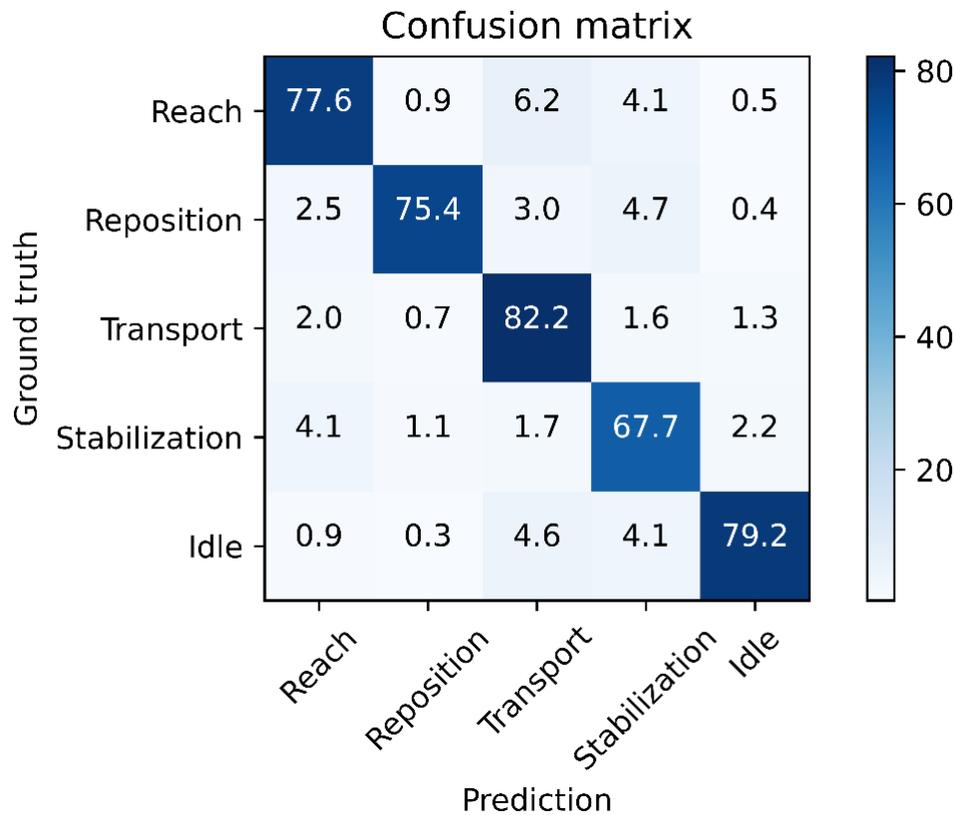

**Supplementary Figure 1. Classification performance of Seq2Seq.** Shown is a confusion matrix, with values normalized to the ground truth primitive count. The diagonal values represent the sensitivity per primitive, or how often the model correct predicted a primitive that was actually performed. The non-diagonal values represent the identification errors made by Seq2Seq. Rows reflect swap-out errors for ground truth primitives and indicate how often a ground truth primitive was incorrectly predicted as another primitive class. Seq2Seq made modest swap-out errors for all primitives (reach, 0.9–6.2%; reposition, 0.4–4.7%; transport, 0.7–2.0%; stabilizations, 1.1–4.1%; idle, 0.3–4.6%). The columns reflect swap-in errors for predicted primitives and indicate how often an incorrect primitive was predicted instead of the ground truth primitive. Seq2Seq made modest swap-in errors for all primitives (reach, 0.9–4.1%; reposition, 0.3–1.1%; transport, 1.7–6.2%; stabilizations, 1.6–4.7%; idle, 0.4–2.2%). We note that some of the errors made by the model could be explained by the lack of finger information from the IMU setup (e.g. confusion between reaches and transports, idles and stabilizations). These primitives have similar motion phenotypes and are distinguished by grasp onset/amount.

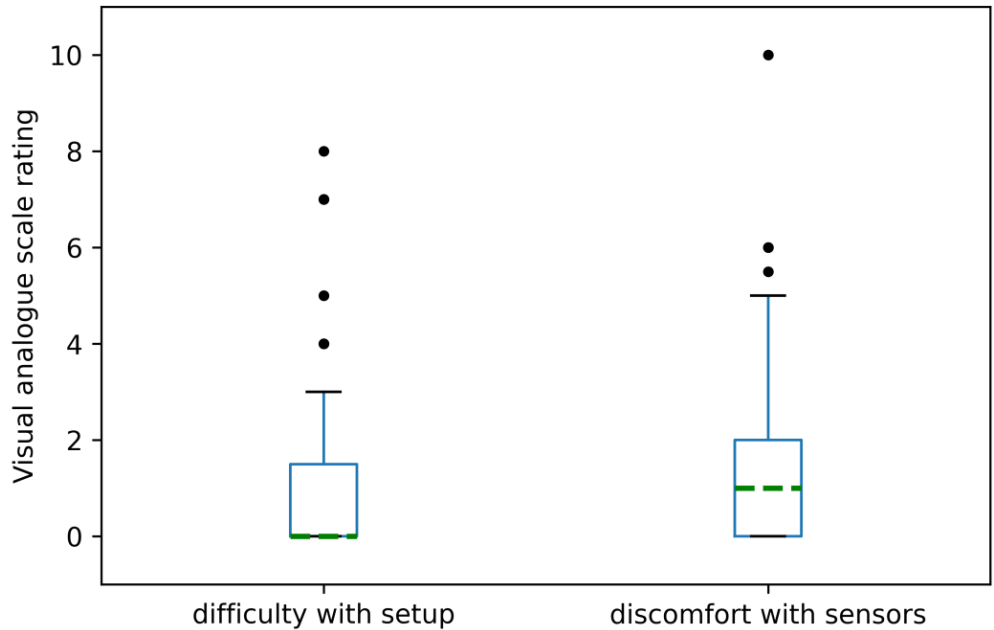

**Supplementary Figure 2. Patient tolerance of IMUs for motion capture.** Shown are box plots of the visual analogue scale (VAS) ratings by the patients. We asked two questions of each patient (n=41) at the end of each session: 'how difficult was the setup (IMU application and calibration)?' and 'how uncomfortable were the IMUs?' Patients scored their responses on an ordinal scale ranging from 0 (not at all) to 10 (most). Most patients reported minimal difficulty with the setup during data collection (median score 0, range 0–8) and minimal discomfort with wearing the IMUs (median score 1, range 0–10), highlighting the unobtrusiveness of IMUs for motion capture. The 25-75[th] interquartile range (IQR) are shown with the lower and upper limits of the box plots, median values and 1.5*IQR are shown with the green dotted line and error bars respectively, and outliers are shown with black dots.

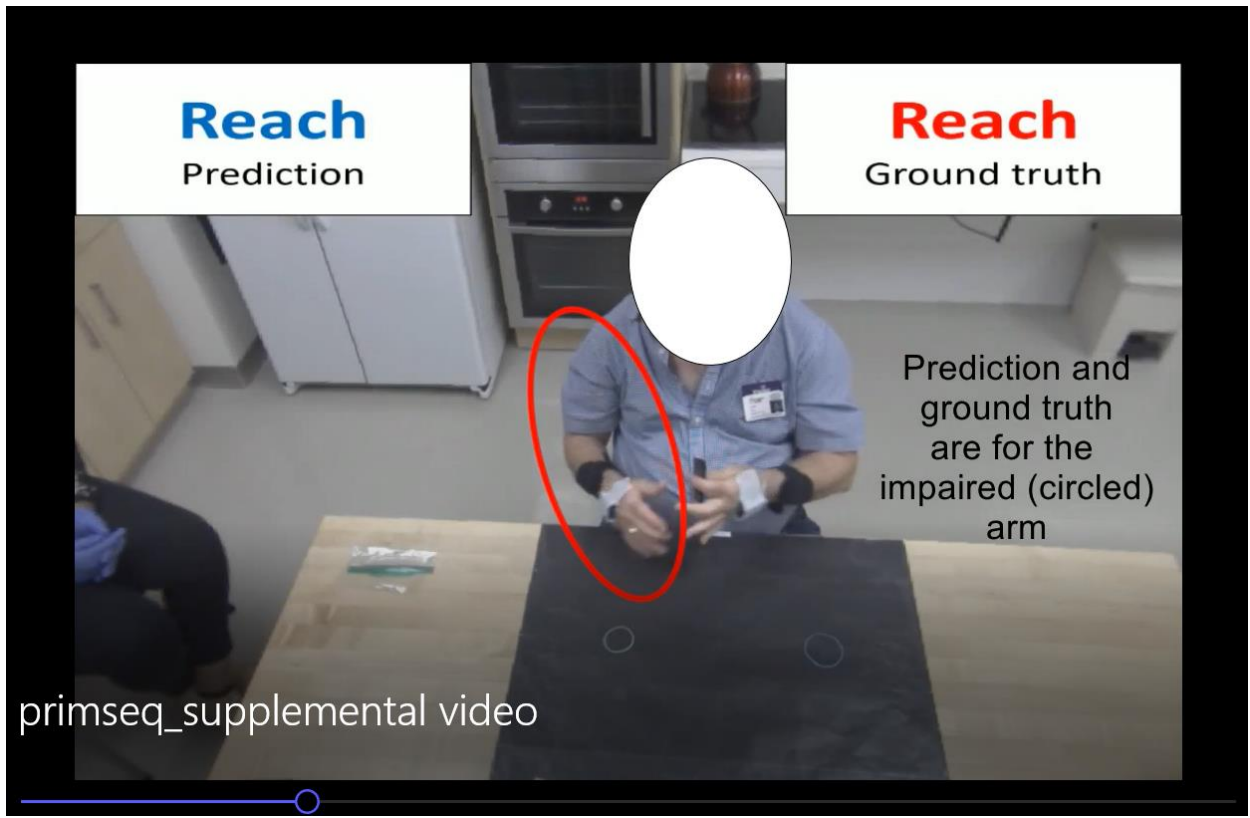

**Supplementary Video. Visualization of model predictions with respect to ground truth primitives.** Shown is a patient performing a combing activity. Human coders used the videotaped activity to identify and label primitives performed by the impaired side (circled); these ground truth labels are shown on the upper right. The trained sequence-to-sequence model used the IMU data to predict primitives performed by the impaired side; these predictions are shown on the upper left.